\documentclass[10pt,twocolumn,letterpaper]{article}

\usepackage{cvpr}
\usepackage{times}
\usepackage{epsfig}
\usepackage{graphicx}
\usepackage{amsmath}
\usepackage{amssymb}
\usepackage{mmstyle}
\usepackage{multirow}
\usepackage{ltablex}
\usepackage{booktabs}
\usepackage{color}
\usepackage{subcaption}
\usepackage{authblk}

\usepackage[pagebackref=true,breaklinks=true,letterpaper=true,colorlinks,bookmarks=false]{hyperref}

\cvprfinalcopy %

\def\cvprPaperID{4252} %

\ifcvprfinal\fi
\begin{document}

\title{FineGym: A Hierarchical Video Dataset for Fine-grained Action Understanding}

\author{Dian Shao \quad Yue Zhao \quad Bo Dai \quad Dahua Lin \\
	CUHK-SenseTime Joint Lab, The Chinese University of Hong Kong\\
	{\tt\small \{sd017, zy317, bdai, dhlin\}@ie.cuhk.edu.hk} 
}

\twocolumn[{
\renewcommand\twocolumn[1][]{#1}
\maketitle
\begin{center}
	\centering
	\vspace{-1em}
	\includegraphics[width=\linewidth]{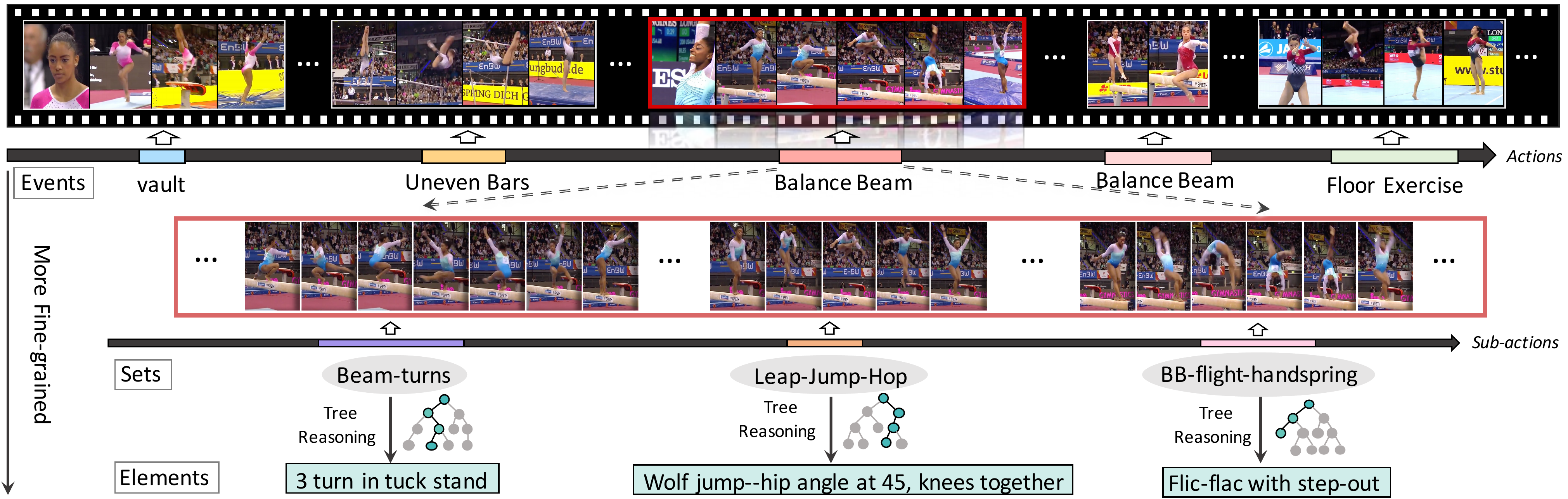}
	\vspace{-2pt}
	\captionof{figure}{An overview of the \emph{FineGym} dataset. We provide coarse-to-fine annotations both temporally and semantically. There are three levels of categorical labels. The temporal dimension (represented by the two bars) is also divided into two levels, \ie, actions and sub-actions. Sub-actions could be described generally using set categories or precisely using element categories. Ground-truth element categories of sub-action instances are obtained via manually constructed decision-trees.}
	\label{fig:overview}
	\vspace{-2pt}
\end{center}
}]
\maketitle
\thispagestyle{empty}
	
\begin{abstract}
\vspace{-9pt}
	On public benchmarks, current action recognition techniques have achieved 
	great success. However, when used in real-world applications, e.g. sport analysis, 
	which requires the capability of parsing an activity into phases and 
	differentiating between subtly different actions, their performances 
	remain far from being satisfactory. 
	To take action recognition to a new level, we develop FineGym\footnote{Dataset and codes can be found at	
	\url{https://sdolivia.github.io/FineGym/}}, 
	a new dataset built on top of gymnastic videos. 
	Compared to existing action recognition datasets, FineGym is distinguished 
	in richness, quality, and diversity. In particular, it provides temporal annotations 
	at both action and sub-action levels with a three-level semantic hierarchy. 
	For example, a ``balance beam'' event will be annotated as a sequence of 
	elementary sub-actions derived from five sets: 
	``leap-jump-hop'', ``beam-turns'', ``flight-salto'', ``flight-handspring'', and ``dismount'', 
	where the sub-action in each set will be further annotated with finely defined class labels. 
	This new level of granularity presents significant challenges for action recognition, 
	\eg~how to parse the temporal structures from a coherent action, and how to distinguish between 
	subtly different action classes. 
	We systematically investigate representative methods on this dataset and 
	obtain a number of interesting findings. We hope this dataset could advance 
	research towards action understanding.
\end{abstract}

\section{Introduction}
\label{sec:intro}

The remarkable progress in action recognition~\cite{simonyan2014two,tran2015learning,UCF101,kuehne2011hmdb, shao2020tapos, yang2020tpn}, particularly
the development of many new recognition models, 
\eg~TSN~\cite{wang2018temporal}, TRN~\cite{zhou2017temporalrelation}, and I3D~\cite{carreira2017quo}, have been largely 
driven by large-scale benchmarks, such as 
ActivityNet~\cite{caba2015activitynet} and Kinetics~\cite{kay2017kinetics}.
On these benchmarks, latest techniques have obtained very high accuracies.

Even so, we found that existing techniques and the datasets that underpin their development
are subject to an important limitation, namely, they focus on  
\emph{coarse-grained} action categories, \eg~\emph{``hockey''} vs. \emph{``gymnastics''}. 
To differentiate between these categories, the background context often plays an important role, 
sometimes even more significant than the action itself. 
However, in certain areas, coarse-grained classification is not enough. 
Take sport analytics for example, it usually requires a detailed comparison between fine-grained
classes, \eg~different moves during a vault. 
For such applications, the capability of fine-grained analysis is needed.
It is worth noting that the \emph{fine-grained} capability here involves two aspects:
1) \emph{temporal:} being able to decompose an action into smaller elements along 
the time axis; 
2) \emph{semantical:} being able to differentiate between sub-classes at the next 
level of the taxonomic hierarchy.

To facilitate the study of fine-grained action understanding, 
we develop \emph{FineGym}, short for \emph{Fine-grained Gymnastics}, 
which is a large-scale high-quality action dataset that provides fine-grained 
annotations. 
Specifically, \emph{FineGym} has several distinguished features:
1) \emph{Multi-level semantic hierarchy.}
All actions are annotated with semantic labels at three levels, 
namely \emph{event}, \emph{set}, and \emph{element}. 
Such a semantic hierarchy provides a solid foundation for 
both coarse- and fine-grained action understanding.
2) \emph{Temporal structure.}
All action instances of interest in each video are identified, and they are 
manually decomposed into sub-actions. These annotated temporal structures 
also provide important support to fine-grained understanding, from another aspect.
3) \emph{High quality.}
All videos in the dataset are high-resolution records of high-level professional competitions. Also, careful quality control is enforced to ensure the accuracy,  
reliability, and consistency of the annotations. 
These aspects together make it a rich dataset for research and a reliable benchmark 
for assessment. 
Moreover, we have summarized a systematic
framework for collecting data and annotations, \eg~labeling via decision trees, which can also be applied to the construction of other datasets with
similar requirements.

Taking advantage of the new exploration space offered by \emph{FineGym},
we conducted a series of empirical studies, with the aim of revealing 
the challenges of fine-grained action understanding. 
Specifically, we tested various action recognition techniques and found that 
their performance on fine-grained recognition is still far from being satisfactory.
In order to provide guidelines for future research, we also revisited 
a number of modeling choices, \eg~the sampling scheme and the input data modalities.
We found that for fine-grained action recognition, 
1) sparsely sampled frames are not sufficient to represent action instances.
2) Motion information plays a significantly important role, rather than visual appearance.
3) Correct modeling of temporal dynamics is crucial.
4) And pre-training on datasets which target for coarse-grained action recognition is not always beneficial.
These observations clearly show the gaps between coarse- and fine-grained action recognition.

Overall, our work contributes to the research of action understanding in 
two different ways:
1) We develop a new dataset \emph{FineGym} for fine-grained action understanding, 
which provides high-quality and fine-grained annotations.
In particular, the annotations are in three semantic levels, namely 
\emph{event}, \emph{set}, and \emph{element}, and two temporal levels, 
namely \emph{action} and \emph{sub-action}.
2) We conduct in-depth studies on top of \emph{FineGym}, which reveal the key 
challenges that arise in the fine-grained setting, which may point to new directions
of future research.

\section{Related Work}
\label{sec:related}
\noindent\textbf{Coarse-grained Datasets for Action Recognition.}
Being the foundation of more sophisticated techniques,
the pursuit of better datasets never stops in the area of action understanding.
Early attempts %
could be traced back to KTH~\cite{schuldt2004recognizing} and Weizmann~\cite{blank2005actions}.
More challenging datasets are proposed subsequently, 
including UCF101~\cite{UCF101}, Kinetics~\cite{carreira2017quo}, ActivityNet~\cite{caba2015activitynet}, Moments in Time~\cite{monfortmoments},
and others \cite{kuehne2011hmdb, THUMOS15, yeung2018every, CabaJHG18, zhao2019hacs, sigurdsson2016hollywood, rodriguez2008action, Jhuang:ICCV:2013, weinzaepfel2016human, mettes2016spot, gu2018ava}.
Some of them also provide annotations beyond category labels, 
ranging from temporal locations \cite{THUMOS15, yeung2018every, CabaJHG18, caba2015activitynet, zhao2019hacs, sigurdsson2016hollywood} 
to spatial-temporal bounding boxes \cite{rodriguez2008action, Jhuang:ICCV:2013, weinzaepfel2016human, mettes2016spot, gu2018ava}.
However, all of these datasets target for coarse-grained action understanding (\eg~\emph{hockey}, \emph{skateboarding}, etc.),
in which the background context often provides distinguishing signals, rather than the actions themselves.
Moreover, as reported in~\cite{wang2018temporal,li2018resound}, sometimes a few frames are sufficient for action recognition on these datasets.

\noindent\textbf{Fine-grained Datasets for Action Recognition.}
There are also attempts towards building datasets for fine-grained action recognition \cite{Damen2018EPICKITCHENS, rohrbach2016recognizing, Goyal_2017, gao2014jhu, kuehne2014language, li2018resound}.
Specifically, both Breakfast~\cite{kuehne2014language} and MPII-Cooking 2~\cite{rohrbach2016recognizing} provides annotations for individual steps of various cooking activities.
In~\cite{kuehne2014language} the coarse actions (\eg \emph{Juice}) are decomposed into action units (\eg cut orange),
and in~\cite{rohrbach2016recognizing} the verb parts are defined to be fine-grained classes (\eg~cut in \emph{cutting onion}).
Something-Something~\cite{Goyal_2017} collects 147 classes of daily human-object interactions, such as \emph{moving something down} and \emph{taking something from somewhere}.
Diving48~\cite{li2018resound} is built on 48 fine-grained diving actions, where the labels are combinations of 4 attributes, \eg~\emph{back+15som+15twis+free}.
Compared to these datasets, our proposed \emph{FineGym} has the following characteristics:
	1) the structure hierarchy is more sophisticated (2 temporal levels and 3 semantic levels), and the number of finest classes is significantly larger (e.g. 530 in FineGym vs. 48 in Breakfast);
	2) the actions in FineGym involve rapid movements and dramatic body deformations, raising new challenges for recognition models;
	3) the annotations are obtained with reference to expert knowledge, where a unified standard is enforced across all classes to avoid ambiguities and inconsistencies.

\noindent\textbf{Methods for Action Recognition.} 
Upon \emph{FineGym} we have empirically studied various state-of-the-art action recognition methods.
These methods could be summarized in three pipelines.
The first pipeline adopts a 2D CNN \cite{simonyan2014two, wang2018temporal, feichtenhofer2016convolutional, donahue2015long} to model per-frame semantics, 
followed by a 1D module to account for temporal aggregation.
Specifically, 
TSN~\cite{wang2018temporal} divides an action instance into multiple segments, representing the instance via a sparse sampling scheme.
An average pooling operation is used to fuse per-frame predictions.
TRN \cite{zhou2017temporalrelation} and TSM \cite{lin2019tsm} respectively replace the pooling operation with a temporal reasoning module and a temporal shifting module.
Alternatively, the second pipeline directly utilizes a 3D CNN \cite{tran2015learning, carreira2017quo, tran2018closer, Wang2018non, diba2017temporal} to jointly capture spatial-temporal semantics,
such as Non-local~\cite{Wang2018non}, C3D \cite{tran2015learning}, and I3D \cite{carreira2017quo}.
Recently, 
an intermediate representation (\eg~human skeleton in \cite{yan2018spatial, cheron2015p, choutas2018potion}) is used by several methods, which could be described as the third pipeline.
Besides action recognition,
other tasks of action understanding,
including action detection and localization~\cite{gaidon2013temporal,xu2017rc3d,zhao2017temporal,hou2017tube,Girdhar2019trans,shao2018find}, 
 action segmentation~\cite{lea2017temporal,ding2018weakly}, and action generation~\cite{li2018flow, sun2019relational}, 
also attract many researchers.

\section{The FineGym Dataset}
 \begin{figure*}
	\centering
	\includegraphics[width=\linewidth]{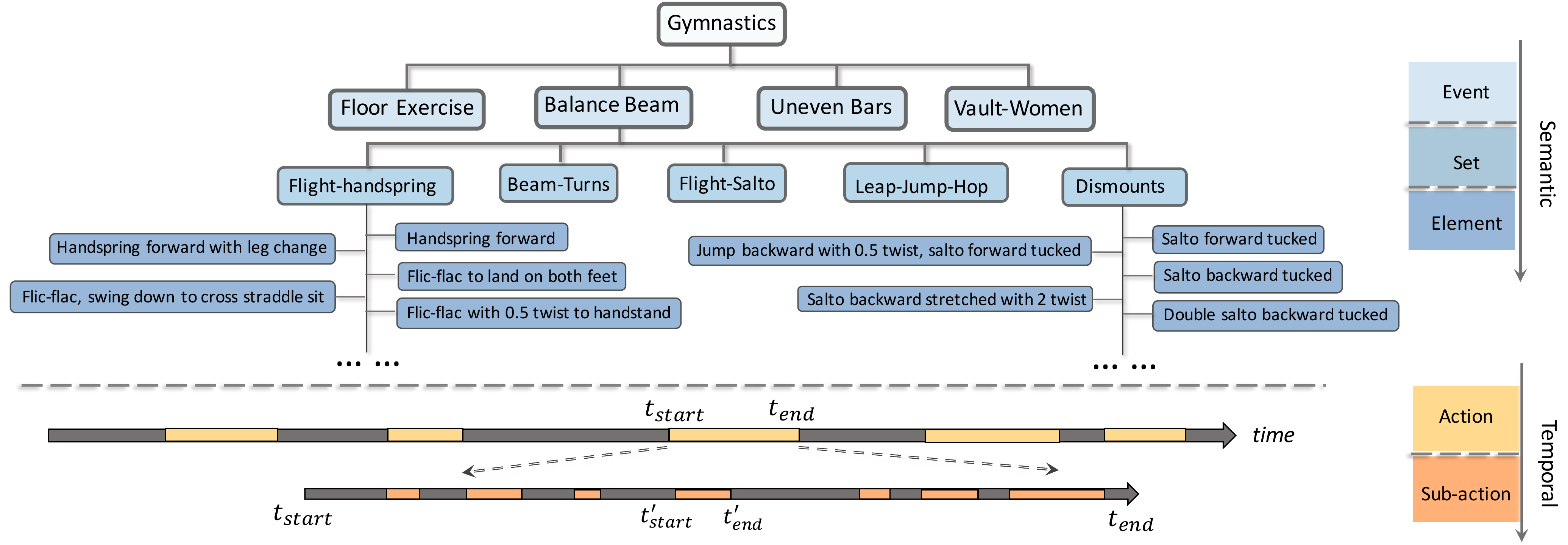}
	\caption{\emph{FineGym} organizes both the semantic and temporal annotations hierarchically. The upper part shows three levels of categorical labels, namely \emph{events} (\eg \emph{balance beam}), \emph{sets} (\eg \emph{dismounts}) and \emph{elements} (\eg \emph{salto forward tucked}). The lower part depicts the two-level temporal annotations, \ie the temporal boundaries of actions (in the top bar) and sub-action instances (in the bottom bar). }
	\label{fig:hierarchy}
\end{figure*}

The goal of our \emph{FineGym} dataset is to introduce a new challenging benchmark with high-quality annotations to the community of action understanding.
While more types of annotations will be included in succeeding versions,
current version of \emph{FineGym} mainly provides annotations for \emph{fine-grained} human action recognition on gymnastics.

Practically, 
categories of actions and sub-actions in \emph{FineGym} are organized according to a three-level hierarchy, 
namely \emph{events}, \emph{sets}, and \emph{elements}.
Events, at the coarsest level of the hierarchy, refer to actions belonging to different gymnastic routines, 
such as \emph{vault} (VT), \textit{floor exercise} (FX), \emph{uneven-bars} (UB), and \emph{balance-beam} (BB).
Sets are mid-level categories, describing sub-actions.
A set holds several technically and visually similar elements.
At the finest granularity are element categories, which equips sub-actions with more detailed descriptions than the set categories.
\eg~a sub-action instance of the set \emph{beam-dismounts} could be more precisely described as \emph{double salto backward tucked} or other element categories in the set.
Meanwhile, \emph{FineGym} also provides two levels of temporal annotations,
namely locations of all events in a video and locations of sub-actions in an action instance (\ie~event instance).
Figure \ref{fig:hierarchy} reveals the annotation organization of \emph{FineGym}.

Below we at first review the key challenges when building \emph{FineGym},
followed by a brief introduction on the construction process,
which covers both data preparation, annotation collection and quality control.
Finally, statistics and properties of \emph{FineGym} are elaborated.

\subsection{Key Challenges}

Building such a complex and fine-grained dataset brings a series of unprecedented challenges,
including:
(1) \emph{How to collect data?} 
Generally, data for large-scale action datasets are mainly collected in two ways,
namely crawling from the Internet and self-recording from invited workers.
However, while fine-grained labels of \emph{FineGym} contain rich details, \eg~\emph{double salto backward tucked with 2 twist},
videos collected in these ways can hardly match the details precisely.
Instead, we collect data from video records of high-level professional competitions.
(2) \emph{How to define and organize the categories? }
With the rich granularities of \emph{FineGym} categories and the subtle differences between instances of the finest categories,
manually defining and organizing \emph{FineGym} categories as in \cite{Goyal_2017, rohrbach2016recognizing} is impractical.
Fortunately, we could resort to official documentation provided by experts \cite{WAG:2017-2020},
which naturally define and organize \emph{FineGym} categories in a consistent way.
This results in $530$ well defined categories.
(3) \emph{How to collect annotations?}
As mentioned, the professional requirements and subtle differences of \emph{FineGym} categories prevent us 
from utilizing crowdsourcing services such as the Amazon Mechanical Turk.
Instead, we hire a team trained specifically for this job.
(4) \emph{How to control the quality?}
Even with a trained team,
the richness and diversity of possible annotations inevitably require an effective and efficient mechanism for quality control,
without which we may face serious troubles as errors would propagate along the hierarchies of \emph{FineGym}.
We thus enforce a series of measures for quality control,
as described in~\ref{subsec:data_collect}.

\subsection{Dataset Construction}
\label{subsec:data_collect}

\noindent\textbf{Data Preparation.}
Our procedure for data collection takes the following steps.
We start by surveying the top-level gymnastics competitions held in recent years.
Then, we collect official video records of them from the Internet, 
ensuring these video records are complete, distinctive and of high-resolutions, \eg 720P and 1080P.
Finally, we cut them evenly into chunks of 10-minutes for further processing.
Through these steps, the quality of data is ensured by the choice of official video records.
The temporal structures of actions and sub-actions are also guaranteed as official competitions are consistent and rich in content.
Moreover, data redundancy is avoided through manual checking.

 \begin{figure}
	\centering
	\includegraphics[width=\linewidth]{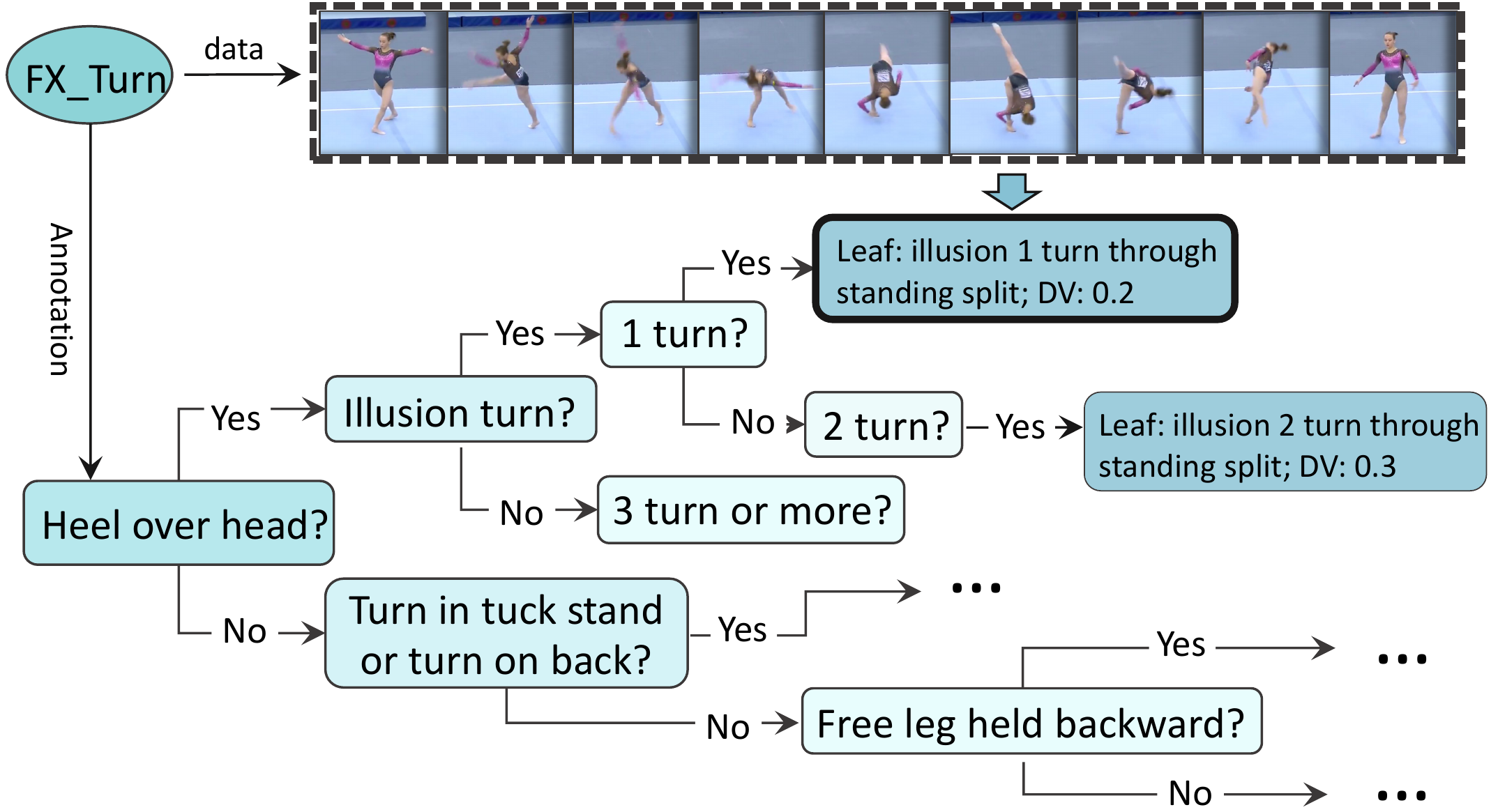}
	\caption{Illustration of the decision-tree based reasoning process for annotating element labels within a given set (\eg \emph{FX-turns}). }
	\label{fig:tree}
\end{figure}

\vspace{2pt}
\noindent\textbf{Annotation Collection.}
We adopt a multi-stage strategy to collect the annotations for both the three-level semantic category hierarchy (\ie~event, set and element labels) 
and the two-level temporal structures of action instances.
The whole annotation process is illustrated in Figure~\ref{fig:overview}, and described as follows: 
1) Firstly, annotators %
are asked to accurately locate the start and end time of each complete gymnastics routine (\ie \textit{an complete action instance containing several sub-actions}) in a video record, and then select the correct event label for it. 
In this step, we discard all incomplete routines, such as the ones that have an interruption.
2) Secondly, 15 sets from 4 events are selected from the latest official codebook~\cite{WAG:2017-2020}, for they provide more distinctive element-level classes.
We further discard the element-level classes that have visually imperceptible differences and unregulated moves. 
Consequently, %
when given an event,
an annotator will locate all the sub-actions from the defined sets and provide their set-level labels.
3) Each sub-action further requires an element label, which is hard to decide directly.
We thus utilize a decision-tree\footnote{Details of the decision-trees are included in the supplemental material.}
 consisting of attribute-based queries to guide the decision.
Starting from the root, which has a set label, an annotator travels on the tree until he meets a leaf node, which has an element label.
See Figure \ref{fig:tree} for a demonstration.

\vspace{2pt}
\noindent\textbf{Quality Control.}
\label{subsec:data_quality}
To build a high-quality dataset which offers clean annotations at all hierarchies,
we adopt a series of mechanisms including: 
training annotators with domain-specific knowledge,
pretesting the annotators rigorously before formal annotation,
preparing referential slides as well as demos,
and cross-validating across annotators.

\begin{table}[]
	\small
	\centering
	\begin{tabular}{c||c|l||c|c}
		\toprule
		event                       & \# set cls              & \# element cls & \# inst & \# sub\_inst \\ 
		\midrule[1pt]
		VT                          & 1                       & 67             & 2034    & 2034         \\ 
		FX                          & 5                       & 64+20+23+4     & 912    & 8929         \\ 
		BB                          & 5                       & 58+25+26+26    & 976    & 11586        \\ 
		UB                          & 4                       & 52+22+57+2     & 961    & 10148        \\  
		\midrule[1pt]
		10 in total						& 15					  & 530			   & 4883    &32697 \\ 
		\bottomrule
	\end{tabular}
	\caption{The statistics of \emph{FineGym} v1.0.}
	\label{tab:data_stat}
\end{table}

 \begin{figure*}
	\centering
	\includegraphics[width=\linewidth]{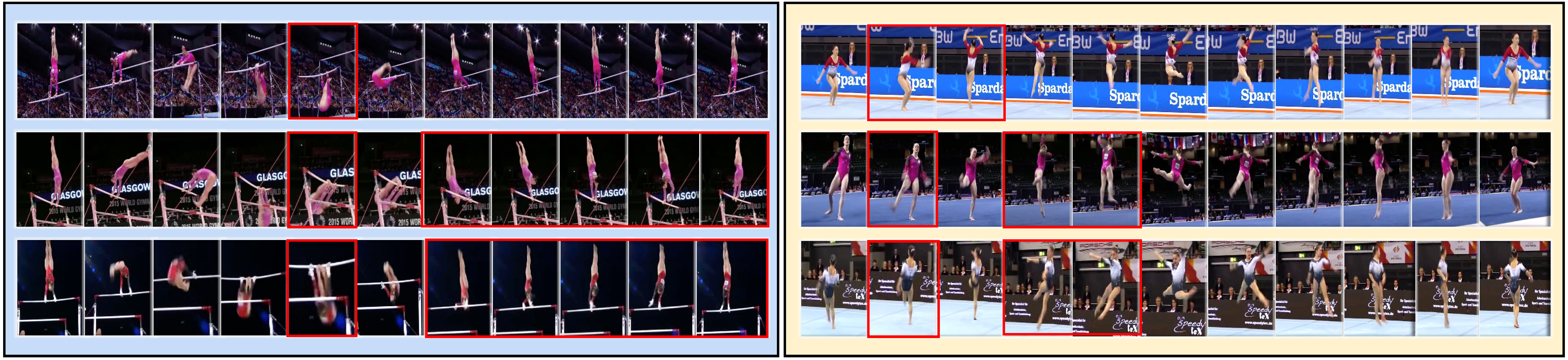}
	\caption{Examples of fine-grained sub-action instances in \emph{FineGym}. 
The left part shows instances belonging to three element categories within the set \emph{UB-circles}, 
from top to bottom: \emph{``clear pike circle backward with $1$ turn"}, 
\emph{``pike sole circle backward with $1$ turn"}, 
and \emph{``pike sole circle backward with $0.5$ turn}.
On the right, there are instances from three element categories of the set \emph{FX-leap-jump-hop}, 
from top to bottom: \emph{``split jump with $1$ turn"}, 
\emph{``split leap with $1$ turn"}, 
and \emph{``switch leap with $1$ turn"}. 
It can be seen such fine-grained instances contain subtle and challenging differences.
Best viewed in high resolution.}
	\label{fig:data_sample}
\end{figure*}

\subsection{Dataset Statistics}
\label{subsec:data_stat}

Table \ref{tab:data_stat} shows the  statistics of \emph{FineGym} v1.0, which is used for empirical studies in this paper.\footnote{More data is provided in the v1.1 release, see the webpage for details.}
Specifically,
\emph{FineGym} contains $10$ event categories, including $6$ male events and $4$ female events.
Particularly, we selected $4$ female events therefrom to provide more fine-grained annotations.
The number of instances in each element category ranges from $1$ to $1,648$, reflecting the natural heavy-tail distribution of them.
$354$ out of the defined $530$ element categories have at least one instance.\footnote{The overall distribution of element categories is presented in the supplementary material.}
To meet different demands, besides the naturally imbalanced setting,
we also provide a more balanced setting by thresholding the number of instances.
Details are included in Sec.\ref{subsec:exp-element}.
In terms of other statistics, 
there are currently $303$ competition records, amounted to $\sim708$ hours. 
For the $4$ events with finer annotations,
\emph{Vault} has a relatively shorter duration ($8$s in average) and intenser motions,
while others have a relatively longer duration ($55$s in average). 
Being temporally more fine-grained, the annotated sub-action instances usually cover less than $2$ seconds,
satisfying the prerequisite to being temporally short for learning more fine-grained information~\cite{Goyal_2017}.

\subsection{Dataset Properties}

\emph{FineGym} has several attracting properties that distinguish it from existing datasets.

\textbf{High Quality.} The videos in \emph{FineGym} are all official recordings of top-level competitions, 
action instances in which are thus professional and standard. 
Besides, over $95\%$ of these videos are of high resolutions (720P and 1080P),
so that subtle differences between action instances are well preserved,
leaving a room for future annotations and models.
Also, due to the utilization of a well-trained annotation team and official documents of category definitions and organizations, 
annotations in \emph{FineGym} are consistent and clean
across different aspects.

\textbf{Richness and Diversity.}
As discussed, \emph{FineGym} contains multiple granularities both semantically and temporally. 
While the number of categories increases significantly when we move downwards along the semantic hierarchy,
the varying dynamics captured in temporal granularities lay a foundation for more comprehensive temporal analysis.
Moreover, \emph{FineGym} is also rich and diverse in terms of viewpoints and poses.
For example, many rare poses are covered in \emph{FineGym} due to actions like twist and salto.

\textbf{Action-centric Instances.}
Unlike several existing datasets where the background is also a major factor for distinguishing different categories,
all instances in \emph{FineGym} have relatively consistent backgrounds.
Moreover, being the same at first glance, 
instances from two different categories may only have subtle differences, especially at the finest semantic granularity.
\eg~the bottom two samples in the right of Figure~\ref{fig:data_sample} differ in whether the directions of legs and the turn are consistent at the beginning.
We thus believe \emph{FineGym} is one of the challenging datasets that requires more focus on the actions themselves. 

\textbf{Decision Trees of Element Categories.}
As we annotate element categories using manually built decision trees consisting of attribute-based queries,
the path from a tree's root to one of its leaf node naturally offers more information than just an element label,
such as the attribute sets and the difficulty score of an element.
Potentially one could use these decision trees for prediction interpretation and reasoning.

\begin{table}[]
	\small
	\centering

	\begin{tabular}{c @{\hspace{1\tabcolsep}}ccccc}
		\toprule[1pt]
		\multicolumn{2}{c}{\textbf{Model Info}}                            & \multicolumn{2}{c}{\textbf{\textit{Event}}}                     & \multicolumn{2}{c}{\textbf{\textit{Set}}}                      \\ 
		\midrule[1pt]
		\textbf{Method}                        & \multicolumn{1}{c}{\textbf{Modality}} & \textbf{Mean} & \multicolumn{1}{c}{\textbf{Top-1}} & \multicolumn{1}{c}{\textbf{Mean}} & \textbf{Top-1} \\ 
		\bottomrule[1pt]

		\multicolumn{1}{c|}  {\multirow{3}{*}{TSN~\cite{wang2018temporal}} }         		 & \multicolumn{1}{c|}{RGB}      		& 98.42       & \multicolumn{1}{c|}{98.18}      & 89.85         & 95.25      \\
		\multicolumn{1}{c|}{}            										& \multicolumn{1}{c|}{Flow}     		& 93.40       & \multicolumn{1}{c|}{93.25}      & 91.64                             & 96.42      \\
		\multicolumn{1}{c|}{}	 		  									    & \multicolumn{1}{c|}{2Stream}  	& 98.47        & \multicolumn{1}{c|}{99.86}      & 91.97                            & 97.69     \\ 
		\toprule[1pt]
	\end{tabular}
\caption{Results of Temporal Segment Network (TSN) in terms of coarse-grained (\ie event and set level) action recognition.}
\label{tab:recog_event_group}

\end{table}

\section{Empirical Studies}
On top of \emph{FineGym}, 
we systematically evaluate representative action recognition methods across multiple granularities, 
and also include a demonstrative study on a typical action localization method using MMAction~\cite{zhao2019mmaction}.
All training protocols follow the original papers unless stated otherwise.
Our main focus is on understanding fine-grained actions (\ie element-level), 
whose challenging characteristics could lead to new inspirations.
Finally, we provide some heuristic observations for future research on this direction.

\begin{table*}[t]
\begin{subtable}[h]{.33\textwidth}
		\scriptsize
	\centering
	\setlength{\tabcolsep}{0.5pt}
	\begin{tabular}{cccccc}
		\toprule[1pt]
		\multicolumn{2}{c}{\textbf{Model Info}}                            & \multicolumn{2}{c}{\textbf{\textit{Gym288}}}                     & \multicolumn{2}{c}{\textbf{\textit{Gym99}}}                      \\ 
		\midrule[1pt]
		\textbf{Method}                        & \multicolumn{1}{c}{\textbf{Modality}} & \textbf{Mean} & \multicolumn{1}{c}{\textbf{Top-1}} & \multicolumn{1}{c}{\textbf{Mean}} & \textbf{Top-1} \\     \bottomrule[1pt]
		\multicolumn{1}{c|}{Random}                  				      & \multicolumn{1}{c|}{-}      		       & 0.3        	  & \multicolumn{1}{c|}{-}              & 1.0                             & --      \\ \hline
		\multicolumn{1}{c|}{ActionVLAD~\cite{girdhar2017actionvlad}}                  				& \multicolumn{1}{c|}{RGB}      		& 16.5        & \multicolumn{1}{c|}{60.5}      & 50.1                             & 69.5      \\ \hline
		\multicolumn{1}{c|}  {\multirow{3}{*}{TSN~\cite{wang2018temporal}} }         		 & \multicolumn{1}{c|}{RGB}      		& 26.5        & \multicolumn{1}{c|}{68.3}      & 61.4                             & 74.8      \\
		\multicolumn{1}{c|}{}            										& \multicolumn{1}{c|}{Flow}     		& 38.7        & \multicolumn{1}{c|}{78.3}      & 75.6                             & 84.7      \\
		\multicolumn{1}{c|}{}	 		  									    & \multicolumn{1}{c|}{2Stream}  	& 37.6        & \multicolumn{1}{c|}{79.9}      & 76.4                             & 86.0      \\ \hline
		\multicolumn{1}{c|}{\multirow{3}{*}{TRN~\cite{zhou2017temporalrelation}}}            		 & \multicolumn{1}{c|}{RGB}      		& 33.1        & \multicolumn{1}{c|}{73.7}      & 68.7                             & 79.9      \\
		\multicolumn{1}{c|}{}													&\multicolumn{1}{c|}{Flow}     			& 42.6        & \multicolumn{1}{c|}{79.5}      & 77.2                             & 85.0      \\
		\multicolumn{1}{c|}{}													& \multicolumn{1}{c|}{2Stream}  	& 42.9        & \multicolumn{1}{c|}{81.6}      & 79.8                             & 87.4      \\ \hline
		\multicolumn{1}{c|}{\multirow{3}{*}{TRNms~\cite{zhou2017temporalrelation}}} 	& \multicolumn{1}{c|}{RGB}      		& 32.0        & \multicolumn{1}{c|}{73.1}      & 68.8                             & 79.5      \\
		\multicolumn{1}{c|}{}													& \multicolumn{1}{c|}{Flow}     		& 43.4        & \multicolumn{1}{c|}{79.7}      & 77.6                             & 85.5      \\
		\multicolumn{1}{c|}{}													& \multicolumn{1}{c|}{2Stream}  	 & 43.3        & \multicolumn{1}{c|}{82.0}      & 80.2                              & 87.8      \\ \hline
		\multicolumn{1}{c|}{\multirow{3}{*}{TSM~\cite{lin2019tsm}}}          			 & \multicolumn{1}{c|}{RGB}      	  & 34.8        & \multicolumn{1}{c|}{73.5}        & 70.6                             & 80.4      \\
		\multicolumn{1}{c|}{}													& \multicolumn{1}{c|}{Flow}     		& 46.0        & \multicolumn{1}{c|}{81.6}      & 80.3                             & 87.1      \\
		\multicolumn{1}{c|}{}													& \multicolumn{1}{c|}{2Stream}  	& 46.5        & \multicolumn{1}{c|}{83.1}      & 81.2                             & 88.4      \\ \hline
		\multicolumn{1}{c|}{I3D~\cite{carreira2017quo}}         		& \multicolumn{1}{c|}{RGB}      & 27.9        & \multicolumn{1}{c|}{66.7}      & 63.2                             & 74.8      \\
		\multicolumn{1}{c|}{I3D$^{*}$~\cite{carreira2017quo}}         		& \multicolumn{1}{c|}{RGB}      & 28.2        & \multicolumn{1}{c|}{66.1}      & 64.4                             & 75.6      \\ \hline
		\multicolumn{1}{c|}{NL I3D~\cite{Wang2018non}}      	& \multicolumn{1}{c|}{RGB}      & 27.1        & \multicolumn{1}{c|}{64.0}      & 62.1                             & 73.0      \\
		\multicolumn{1}{c|}{NL I3D$^{*}$~\cite{Wang2018non}}      	  & \multicolumn{1}{c|}{RGB}      & 28.0        & \multicolumn{1}{c|}{67.0}      & 64.3                             & 75.3      \\ \hline
		\multicolumn{1}{c|}{ST-GCN~\cite{yan2018spatial}}                         			& \multicolumn{1}{c|}{Pose}     &   11.0        & \multicolumn{1}{c|}{34.0}           & 25.2                                   &36.4            \\  \toprule[1pt]
	\end{tabular}
	\caption{Results of \emph{elements across all events}.}
	\label{tab:recog_element_all}
\end{subtable}%
\begin{subtable}[h]{.33\textwidth}
	\scriptsize
\centering
\setlength{\tabcolsep}{0.5pt}
\begin{tabular}{cccccc}
	\toprule[1pt]
	\multicolumn{2}{c}{\textbf{Model Info}}                            & \multicolumn{2}{c}{\textbf{VT}, \textit{6cls}}                     & \multicolumn{2}{c}{\textbf{FX}, \textit{35cls}}                      \\ 
	\midrule[1pt]
	\textbf{Method}                        & \multicolumn{1}{c}{\textbf{Modality}} & \textbf{Mean} & \multicolumn{1}{c}{\textbf{Top-1}} & \multicolumn{1}{c}{\textbf{Mean}} & \textbf{Top-1} \\   \bottomrule[1pt]
	\multicolumn{1}{c|}{Random}                  				      & \multicolumn{1}{c|}{-}      		       & 16.7        	  & \multicolumn{1}{c|}{-}              & 2.9                             & --      \\ \hline
	\multicolumn{1}{c|}{ActionVLAD~\cite{girdhar2017actionvlad}}                  				& \multicolumn{1}{c|}{RGB}      		  & 32.7        	& \multicolumn{1}{c|}{44.6}      & 56.4                             & 65.0      \\ \hline
	\multicolumn{1}{c|}  {\multirow{3}{*}{TSN~\cite{wang2018temporal}} }         		 & \multicolumn{1}{c|}{RGB}      		   & 27.8         & \multicolumn{1}{c|}{46.6}       & 58.6                             & 67.5     \\
	\multicolumn{1}{c|}{}            										& \multicolumn{1}{c|}{Flow}     		   & 23.1         & \multicolumn{1}{c|}{42.6}      & 70.7                             & 78.5      \\
	\multicolumn{1}{c|}{}	 		  									    & \multicolumn{1}{c|}{2Stream}  	    & 27.0         & \multicolumn{1}{c|}{47.5}       & 73.1                             & 81.6      \\ \hline
	\multicolumn{1}{c|}{\multirow{3}{*}{TRN~\cite{zhou2017temporalrelation}}}            		 & \multicolumn{1}{c|}{RGB}      		    &32.1        & \multicolumn{1}{c|}{48.0 }       &65.8                           & 72.0      \\
	\multicolumn{1}{c|}{}													&\multicolumn{1}{c|}{Flow}     			   & 28.9         & \multicolumn{1}{c|}{44.2}       & 74.9                             & 81.2     \\
	\multicolumn{1}{c|}{}													& \multicolumn{1}{c|}{2Stream}  	    & 31.4         & \multicolumn{1}{c|}{47.1}       & 77.5                             & 84.6    \\ \hline
	\multicolumn{1}{c|}{\multirow{3}{*}{TRNms~\cite{zhou2017temporalrelation}}} 	& \multicolumn{1}{c|}{RGB}      		  & 31.5         & \multicolumn{1}{c|}{46.6}        & 66.6                           & 73.4      \\
	\multicolumn{1}{c|}{}													& \multicolumn{1}{c|}{Flow}     		   & 29.1         & \multicolumn{1}{c|}{43.9}        & 74.8                           & 81.1      \\
	\multicolumn{1}{c|}{}													& \multicolumn{1}{c|}{2Stream}  	    & 30.1         & \multicolumn{1}{c|}{47.3}        & 78.2                             & 84.9     \\ \hline
	\multicolumn{1}{c|}{\multirow{3}{*}{TSM~\cite{lin2019tsm}}}          			 & \multicolumn{1}{c|}{RGB}      		   & 29.2        & \multicolumn{1}{c|}{42.2}        & 62.2                             & 68.8      \\
	\multicolumn{1}{c|}{}													& \multicolumn{1}{c|}{Flow}     		   & 26.2         & \multicolumn{1}{c|}{42.4}        & 76.2                             & 81.9      \\
	\multicolumn{1}{c|}{}													& \multicolumn{1}{c|}{2Stream}  		& 28.8         & \multicolumn{1}{c|}{44.8}        & 76.9                             & 83.6      \\ \hline
	\multicolumn{1}{c|}{{I3D}~\cite{carreira2017quo}}         	   & \multicolumn{1}{c|}{RGB}      			  & 31.5         & \multicolumn{1}{c|}{42.1}        & 53.7                             & 59.5      \\
	\multicolumn{1}{c|}{I3D$^{*}$~\cite{carreira2017quo}}         		 & \multicolumn{1}{c|}{RGB}      		 	& 33.4         & \multicolumn{1}{c|}{47.8}        & 52.2                             & 60.2     \\ \hline
	\multicolumn{1}{c|}{NL I3D~\cite{Wang2018non}}      	& \multicolumn{1}{c|}{RGB}      		   & 30.6          & \multicolumn{1}{c|}{46.0}        & 53.4                             & 59.8     \\
	\multicolumn{1}{c|}{NL I3D$^{*}$~\cite{WAG:2017-2020}}      	  & \multicolumn{1}{c|}{RGB}      		  	  & 30.8         & \multicolumn{1}{c|}{47.3}         & 50.9                             & 57.6      \\ \hline
	\multicolumn{1}{c|}{ST-GCN~\cite{yan2018spatial}}                         			& \multicolumn{1}{c|}{Pose}     		    &     19.5          	&  \multicolumn{1}{c|}{38.8}                  &    35.3                                  &   40.1         \\     \toprule[1pt]
\end{tabular}
\caption{Results of \emph{elements within a event}.}
\label{tab:recog_element_event}
\end{subtable}%
\begin{subtable}[h]{.33\textwidth}
	\scriptsize
\centering
\setlength{\tabcolsep}{0.5pt}
\begin{tabular}{cccccc}
	\toprule[1pt]
	\multicolumn{2}{c}{\textbf{Model Info}}                            & \multicolumn{2}{c}{\textbf{FX-S1}, \textit{11cls}}                     & \multicolumn{2}{c}{\textbf{UB-S1}, \textit{15cls}}                      \\ 
	\midrule[1pt]
	\textbf{Method}                        & \multicolumn{1}{c}{\textbf{modality}} & \textbf{Mean} & \multicolumn{1}{c}{\textbf{Top-1}} & \multicolumn{1}{c}{\textbf{Mean}} & \textbf{Top-1} \\     \bottomrule[1pt]
	\multicolumn{1}{c|}{Random}                  				& \multicolumn{1}{c|}{-}      		               & 9.1        	& \multicolumn{1}{c|}{-}               & 6.7                             & --      \\ \hline
	\multicolumn{1}{c|}{ActionVLAD~\cite{girdhar2017actionvlad}}                  				& \multicolumn{1}{c|}{RGB}      		  & 45.0        	& \multicolumn{1}{c|}{52.3}      & 51.9                             & 64.6      \\ \hline
	\multicolumn{1}{c|}  {\multirow{3}{*}{TSN~\cite{wang2018temporal}} }         		 & \multicolumn{1}{c|}{RGB}      		   & 31.2         & \multicolumn{1}{c|}{49.9}       & 44.8                             & 65.6      \\
	\multicolumn{1}{c|}{}            										& \multicolumn{1}{c|}{Flow}        & 69.6         & \multicolumn{1}{c|}{78.0}      & 65.3                             & 78.9      \\
	\multicolumn{1}{c|}{}	 		  									    & \multicolumn{1}{c|}{2Stream}  	 & 68.2        & \multicolumn{1}{c|}{78.5}       & 65.0                             & 80.0      \\ \hline
	\multicolumn{1}{c|}{\multirow{3}{*}{TRN~\cite{zhou2017temporalrelation}}}            		 & \multicolumn{1}{c|}{RGB}      		    & 58.2        & \multicolumn{1}{c|}{55.0 }       &53.6                            & 70.9      \\
	\multicolumn{1}{c|}{}													&\multicolumn{1}{c|}{Flow}     			   & 73.3         & \multicolumn{1}{c|}{79.9}       & 71.5                             & 82.5     \\
	\multicolumn{1}{c|}{}													& \multicolumn{1}{c|}{2Stream}  	    & 74.4         & \multicolumn{1}{c|}{81.9}       & 83.0                             & 71.0      \\ \hline
	\multicolumn{1}{c|}{\multirow{3}{*}{TRNms~\cite{zhou2017temporalrelation}}} 	& \multicolumn{1}{c|}{RGB}      		  & 58.5         & \multicolumn{1}{c|}{64.4}        & 55.8                             & 71.4     \\
	\multicolumn{1}{c|}{}													& \multicolumn{1}{c|}{Flow}        & 75.8         & \multicolumn{1}{c|}{82.6}        & 70.8                            & 82.2      \\
	\multicolumn{1}{c|}{}													& \multicolumn{1}{c|}{2Stream}  	    & 72.9         & \multicolumn{1}{c|}{80.8}        & 70.8                             &83.2      \\ \hline
	\multicolumn{1}{c|}{\multirow{3}{*}{TSM~\cite{lin2019tsm}}}          			 & \multicolumn{1}{c|}{RGB}      		   & 45.6        & \multicolumn{1}{c|}{53.3}        & 50.9                            & 66.4     \\
	\multicolumn{1}{c|}{}													& \multicolumn{1}{c|}{Flow}     		   & 75.8         & \multicolumn{1}{c|}{81.7}        & 73.1                             & 82.5     \\
	\multicolumn{1}{c|}{}													& \multicolumn{1}{c|}{2Stream}  		& 72.9         & \multicolumn{1}{c|}{79.4}        & 70.1                             & 80.8      \\ \hline
	\multicolumn{1}{c|}{I3D~\cite{carreira2017quo}}         	   & \multicolumn{1}{c|}{RGB}      			  & 33.3         & \multicolumn{1}{c|}{38.9}        & 32.2                             & 49.1      \\
	\multicolumn{1}{c|}{I3D$^{*}$~\cite{carreira2017quo}}         		 & \multicolumn{1}{c|}{RGB}      		 	& 36.1         & \multicolumn{1}{c|}{42.9}        & 31.0                             & 48.1     \\ \hline
	\multicolumn{1}{c|}{NL I3D~\cite{Wang2018non}}      	& \multicolumn{1}{c|}{RGB}      		   &31.4         & \multicolumn{1}{c|}{39.0}        & 29.3                             & 48.5     \\
	\multicolumn{1}{c|}{NL I3D$^{*}$~\cite{Wang2018non}}      	  & \multicolumn{1}{c|}{RGB}      		  	  & 35.8        & \multicolumn{1}{c|}{40.1}         & 26.9                           & 48.5      \\ \hline
	\multicolumn{1}{c|}{ST-GCN~\cite{yan2018spatial}}                         			& \multicolumn{1}{c|}{Pose}     		    &  21.6             	& \multicolumn{1}{c|}{30.8}                  &   13.7                                   & 28.1           \\   \toprule[1pt]
\end{tabular}
\caption{Results of \emph{elements within a set}.}
\label{tab:recog_element_group}
\end{subtable}
\caption{Element-level action recognition results of representative methods. 
Specifically, results of recognizing element categories across all events, within an event, and within a set, are respectively included in (a), (b), and (c).}
\label{tab:recog_element}
\end{table*}

\subsection{Event-/Set-level Action Recognition}

We present a brief demonstrative study for the event and set level action recognition,
as their characteristics resemble the coarse-grained action recognition that is well studied in multiple benchmarks.
Specifically, 
we choose the widely adopted Temporal Segment Networks (TSN) \cite{wang2018temporal} as the representative.
It divides an instance into $3$ segments and samples one frame from each segment to form the input.
Visual appearance (RGB) and motion (Optical Flow) features of the input frames are separately processed in TSN,
making it a good choice for comparing the contribution of each feature source.
The results of event and set level action recognition are listed in Table~\ref{tab:recog_event_group},
from which we observe: 
1) $3$ frames, accounting for less than $5\%$ of all frames, are sufficient for recognizing event and set categories,
suggesting categories at these two levels %
could be well classified using isolated frames.
2) Compared to motion features, appearance features contribute more at the event-level, and vice versa at the set level.
This means the reliance on static visual cues such as background context is decreased as we step into a finer granularity.
Such trend continues and becomes clearer at the finest granularity, as shown in the element-level action recognition.

 \begin{figure*}
	\centering
	\includegraphics[width=\linewidth]{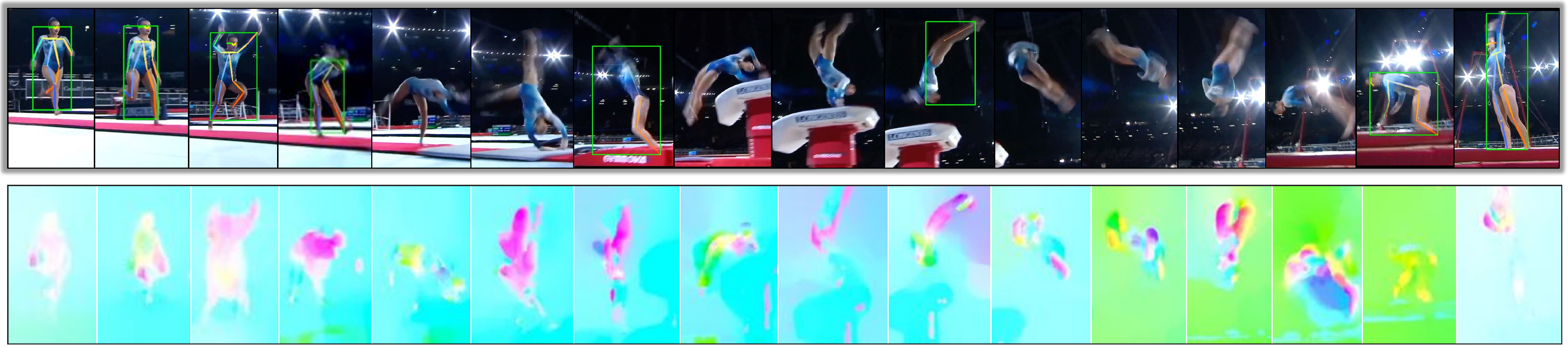}
	\caption{\textbf{Top-row} represents the results of person detection and pose estimation~\cite{fang2017rmpe,li2018crowdpose} for a \emph{Vault} routine, 
and the \textbf{bottom-row} visualizes the optical flow~\cite{zach2007duality} features. 
It can be seen that detections and pose estimations of the gymnast are missed in multiple frames, especially in frames with intense motion.
Best viewed in high resolution.}
\label{fig:pose}
\end{figure*}

\subsection{Element-level Action Recognition}
\label{subsec:exp-element}

We mainly focus on the element-level action recognition,
which raises significant challenges for existing methods.
Specifically, representative methods belonging to various pipelines are selected,
including 2D-CNN (\ie~TSN~\cite{wang2018temporal}, TRN~\cite{zhou2017temporalrelation}, TSM~\cite{lin2019tsm}, and ActionVLAD~\cite{girdhar2017actionvlad}),
3D-CNN methods (\ie~I3D~\cite{carreira2017quo}, Non-local~\cite{Wang2018non}),
as well as a skeleton-based method ST-GCN \cite{yan2018spatial}.

These methods are thoroughly studied in three sub-tasks,
namely recognition of \emph{elements across all events}, \emph{elements within an event}, and \emph{elements within a set},
as shown in Figure \ref{tab:recog_element} (a), (b), and (c) respectively.
For \emph{elements across all events}, we adopt both a natural long-tailed setting and a more balanced setting,
respectively referred to as \emph{Gym288} and \emph{Gym99}.
Details of these settings are included in the supplemental material.
For \emph{elements within an event}, we separately select from \emph{Gym99} all elements of two specific events, namely \emph{Vault (VT)} and \emph{Floor Exercise (FX)}.
The elements of \emph{FX} come from $4$ different sets, 
while the elements of \emph{VT} come from a single set (\emph{VT} is a special event with only one set).
Finally for \emph{elements within a set}, 
we select the set \emph{FX-G1} covering leaps, jumps and hops of \emph{FX},
and the set \emph{UB-G1} covering circles in \emph{Uneven Bars (UB)}.

From the results of these tasks in Table~\ref{tab:recog_element},
we have summarized several observations.
(1) Given the long-tail nature of the instance distribution, all methods are shown to overfit to the elements having the most number of instances,
especially on the setting \emph{Gym288}.
(2) Due to the subtle differences between elements, visual appearances in the form of RGB values 
contribute significantly less than that in coarse-grained action recognition.
And motion features contribute a lot in most cases except for the \emph{Vault} in \emph{elements within an event},
for motion dynamics of elements in \emph{Vault} are very intense.
(3) Capturing temporal dynamics is important as TRN and TSM outperform TSN by large margins.
(4) I3D and Non-local network pre-trained on ImageNet and Kinetics obtain similar results with 2D-CNN methods,
which may be due to the large gap between temporal patterns of element categories and those from Kinetics.
(5) Skeleton-based ST-GCN struggles due to the challenges in skeleton estimation on gymnastics instances,
as shown in %
Figure~\ref{fig:pose}.

\begin{table}[]
\centering
\small
\begin{tabular}{ccccccc}
	\toprule[1pt]
	\multicolumn{7}{c}{\textbf{\emph{GymFine},} \textbf{mAP@}$\alpha$}                                                                             \\ 
	 \bottomrule[1pt]
	\multicolumn{1}{c|}{Temporal level} & 0.50 & 0.60 & \multicolumn{1}{c}{0.70} & 0.80 & \multicolumn{1}{c|}{0.90} & Avg  \\ \bottomrule[1pt]
	\multicolumn{1}{c|}{Action}         & 60.0 & 57.9 & 57.1                      & 54.6 & \multicolumn{1}{c|}{35.0} & 49.4 \\ \hline
	\multicolumn{1}{c|}{Sub-action}     & 22.2 & 15.4 & 9.2                     & 3.9 & \multicolumn{1}{c|}{0.6} & 9.6 \\ \toprule[1pt]
\end{tabular}
\caption{Temporal action localization results of SSN \cite{zhao2017temporal} at coarse- (\ie actions) and fine-grained (\ie sub-actions) levels. 
The metric is mAP@tIoU. 
The average (Avg) values are obtained by ranging tIoU thresholds from 0.5 to 0.95 with an interval of 0.05.}
\label{tab:detection}
\end{table}

\begin{table}[]
\centering
\small
\begin{tabular}{cccc}
	\toprule[1pt]
	\# \textbf{Frame}                 & \textbf{\textit{Gym99} }& \textbf{UCF101~\cite{UCF101}}& \textbf{ActivityNet v1.2~\cite{caba2015activitynet}} \\ 
	\bottomrule[1pt]
	\multicolumn{1}{c|}{1}  & \multicolumn{1}{c|}{35.46} & \multicolumn{1}{c|}{85.0} & 82.0             \\
	\multicolumn{1}{c|}{3}  & \multicolumn{1}{c|}{61.4}  & \multicolumn{1}{c|}{86.5} & 83.6             \\
	\multicolumn{1}{c|}{5}  & \multicolumn{1}{c|}{70.8}  & \multicolumn{1}{c|}{\textbf{86.7}} & \textbf{84.6}             \\
	\multicolumn{1}{c|}{7}  & \multicolumn{1}{c|}{74.4}  & \multicolumn{1}{c|}{86.4} & 84.0             \\
	\multicolumn{1}{c|}{12} & \multicolumn{1}{c|}{\textbf{78.82}} & \multicolumn{1}{c|}{-}    & -                \\ \toprule[1pt]
\end{tabular}
\caption{Performances of TSN when varying the number of sampled frames during training.}
\label{tab:analy_num_seg}
\vspace{-10pt}
\end{table}

\begin{figure*}[]
	\centering
	\begin{subfigure}[t]{0.22\textwidth}
		\includegraphics[height=1.2in]{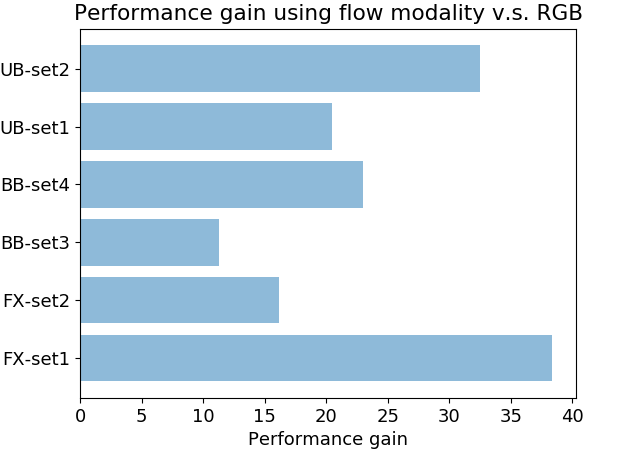}
		\caption{}
		\label{fig:flow_gain}
	\end{subfigure}
    \hspace{0pt}
	\begin{subfigure}[t]{0.22\textwidth}
		\includegraphics[height=1.21in]{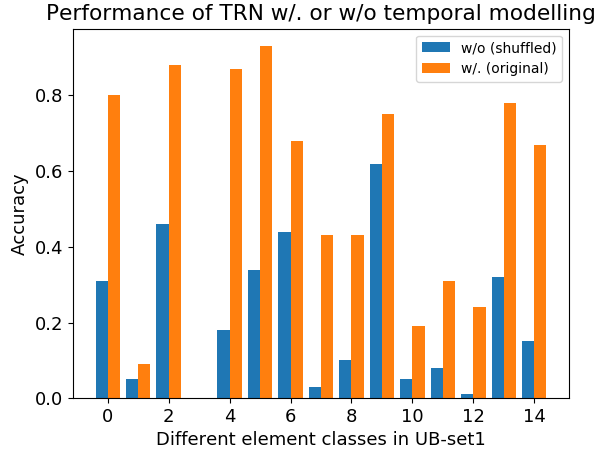}
		\caption{}
		\label{fig:trn_shuffled}
	\end{subfigure}%
	\hspace{8pt}
	\begin{subfigure}[t]{0.22\textwidth}
		\includegraphics[height=1.21in]{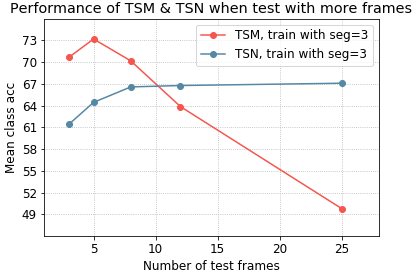}
		\caption{}
		\label{fig:tsm_curve}
	\end{subfigure}
	\hspace{18pt}
	\begin{subfigure}[t]{0.22\textwidth}
	\includegraphics[height=1.28in]{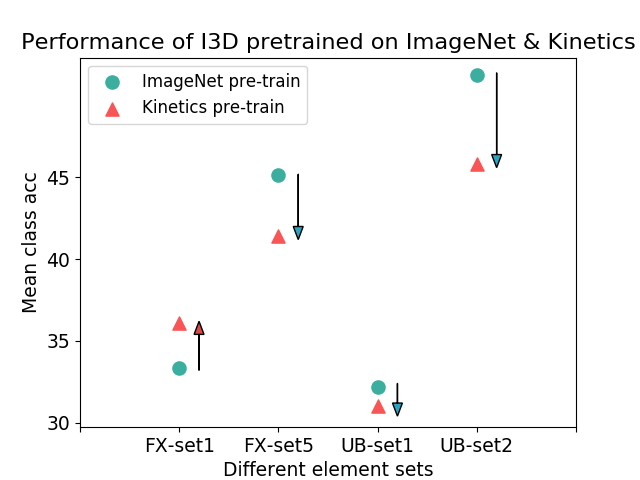}
	\caption{}
	\label{fig:pretrain}
	\end{subfigure}
    \label{fig:analysis}
	\caption{(a) Per-class performances of TSN with motion and appearance features in 6 element categories.
(b) Performances of TRN on the set \emph{UB-circles} using ordered or shuffled testing frames. N
(c) Mean-class accuracies of TSM and TSN on \textit{Gym99} when trained with 3 frames and tested with more frames. 
(d) Per-class performances of I3D pre-trained on Kinetics and ImageNet in various element categories. Best viewed in high resolution.}
\vspace{-7pt}
\end{figure*}

\subsection{Temporal Action Localization}
We also include an illustrative study for temporal action localization,
as \emph{FineGym} could support a wide range of tasks.
Practically, temporal action localization could be conducted for event actions within video records or sub-actions within action instances,
resulting in two sub-tasks.
We select Structured Segment Network (SSN)~\cite{zhao2017temporal} as the representative, relying on its open-sourced implementation.
The results of SSN on these two tasks are listed in Table \ref{tab:detection},
where localizing sub-actions is shown to be much more challenging than localizing actions.
While the boundaries of actions in a video record are more distinctive,
identifying the boundaries of sub-actions may require a comprehensive understanding of the whole action.

\subsection{Analysis}
\label{subsec:analysis}

In this section, we enumerate the key messages we have observed in the conducted empirical studies.

\textbf{Is sparse sampling sufficient for action recognition?}
The sparse sampling scheme has been widely adopted in action recognition, %
due to its high efficiency and promising accuracy demonstrated in various datasets \cite{UCF101, caba2015activitynet}.
However, this trend does not hold for element-level action recognition in \emph{FineGym}.
Table~\ref{tab:analy_num_seg} lists the results of TSN \cite{wang2018temporal} on the subset \emph{Gym99} as well as existing datasets, where we adjust the number of input frames.
Compared to saturated results on existing datasets using only %
few frames, the result of TSN on \emph{Gym99} steadily increases as the number of frames increases,
and saturates at $12$ frames which account for $30\%$ of all frames.
These results indicate that every frame counts in fine-grained action recognition on \emph{FineGym}.

\textbf{How important is temporal information?}
As shown in Figure~\ref{fig:flow_gain}, motion features such as optical flows could capture frame-wise temporal dynamics, 
leading to better performance of TSN \cite{wang2018temporal}.
Many methods have also designed innovative modules for longer-term temporal modeling, such as TRN~\cite{zhou2017temporalrelation} and TSM~\cite{lin2019tsm}.
To study them,
for the temporal reasoning module in TRN, we shuffle the input frames during testing, and observe significant performance drops in Figure \ref{fig:trn_shuffled},
indicating temporal dynamics indeed play an important role in \emph{FineGym}, and TRN could capture it.
Moreover,
for the temporal shifting module in TSM, we conduct a scheme where we start by training a TSM with $3$ input frames,
then gradually increase the number of frames during testing.
Taking TSN for a comparison, Figure~\ref{fig:tsm_curve} includes the resulting curves,
where the performance of TSM drops sharply when the number of testing frames is very different from that in training,
and TSN maintains its performance as only temporal average pooling is applied in it.
These results again verify that temporal dynamics is essential on \emph{FineGym},
so that a very different number of frames leads to significantly different temporal dynamics.
To summarize, optical flows could capture some extent of temporal dynamics, but not all.
Fine-grained action recognition of motion-intense actions heavily relies on temporal dynamics modeling.

\textbf{Does pre-training on large-scale video datasets help?}
Considering the number of parameters in 3D-CNN methods, \eg~I3D \cite{carreira2017quo}, 
usually they are pre-trained first on large-scale datasets, \eg~Kinetics,
which indeed leads to a performance boost \cite{carreira2017quo, zhao2019hacs}.
For example, the Kinetics pre-trained I3D could promote the recognition accuracy from $84.5\%$ to $97.9\%$ on UCF101 \cite{UCF101}.
However, on \emph{FineGym}, such a pre-training scheme is not always helpful, as shown in Figure~\ref{fig:pretrain}.
One potential reason is the large gaps in terms of temporal patterns between coarse- and fine-grained actions.

\textbf{What can not be handled by current methods/modules?}
By carefully observing the confusion matrices~\footnote{See supplementary material for examples.}, we summarize some points that are challenging for existing methods.
(1) \emph{Intense motion}, especially in different kinds of saltos (often finished within $1$ second), as shown in the last several frames of Figure~\ref{fig:pose}.
(2) \emph{Subtle spatial semantics}, which involves differences in body parts such as legs are whether bent or straight, and human-object relationships.
(3) \emph{Complex temporal dynamics}, such as the direction of motion, and the degree of rotation.
(4) \emph{Reasoning}, such as counting the times of saltos.
We hope the high-resolution and professional data of \emph{FineGym} could help future researches aiming for these points.
In addition, \emph{FineGym} poses higher requirements for methods that have an intermediate representation, \eg~human skeleton,
which is hard to be estimated on \emph{FineGym} due to the large diversity in human poses. 
See Figure~\ref{fig:pose} for a demonstration.

\section{Potential Applications and Discussion}

While more types of annotations will be added subsequently,
the high-quality data of \emph{FineGym} has offered a foundation for various applications,
besides coarse- and fine-grained action recognition and localization,
including but not limited to 
(1) \emph{auto-scoring}, where difficult scores are given for each element category in the official documents,
and we could also estimate the quality scores based on visual information,
resulting in a gymnastic auto-scoring framework.
(2) \emph{Action generation}, where the consistent background context of fine-grained sub-actions could help generative models focus more on the action themselves,
and the standard and diverse instances in \emph{FineGym} could facilitate exploration.
(3) \emph{Multi-attribute prediction}, 
for which the attribute ground-truths of the element categories are immediately ready due to the use of decision trees.
(4) \emph{Model interpretation and reasoning}, which could benefit from the manually built decision trees, as shown in Figure~\ref{fig:tree}.

\emph{FineGym} may be used to conduct more empirical studies on model designs,
such as \emph{how to strike a balance between accuracy and efficiency when dealing with highly informative yet subtly different actions?}
And \emph{how to model the complex temporal dynamics efficiently, effectively and robustly?}

\section{Conclusion}
In this paper, we propose \emph{FineGym}, a dataset focusing on gymnastic videos.
\emph{FineGym} differs from existing action recognition datasets in multiple aspects,
including the high-quality and action-centric data, the %
consistent annotations across multiple granularities both semantically and temporally,
as well as the diverse and informative action instances.
On top of \emph{FineGym}, we have empirically investigated representative methods at various levels.
These studies not only lead to a number of attractive findings that are beneficial for future research,
but also clearly show new challenges posed by \emph{FineGym}.
We hope these efforts could facilitate new advances in the field of action understanding.

\vspace{-12pt}
\paragraph{Acknowledgements.}
We sincerely thank the outstanding annotation team for their excellent work.
This work is partially supported by SenseTime Collaborative Grant on Large-scale Multi-modality Analysis and the General Research Funds (GRF) of Hong Kong (No. 14203518
	 and No. 14205719).

{\small
\bibliographystyle{ieee_fullname}
\bibliography{egbib}
}

\end{document}